\documentclass[letterpaper,pdflatex]{article}
\usepackage{times}
\usepackage{courier}
\usepackage{amsmath}
\usepackage{graphicx}
\usepackage{url}

\usepackage{aaai}
\usepackage{clrscode3e}

\def\mvp{\vspace{-0.05in}}

\long\def\Ignore#1{}
\def\acronym#1{\text{\textsc\textbf{\small #1}}}

\newcommand{\proofend}{\ \rule{0.5em}{0.5em}}

\newcommand{\aSet}[1]{{\left\{#1\right\}}}

\def\grad{\nabla}

\def\calG{\mathcal{G}}

\def\Begin{\Indentmore}
\def\Then{\kw{then} \Begin}

\def\CRIKEY3{\acronym{CRIKEY3}}

\def\eps{\varepsilon}

\newcommand{\citep}[1]{\citeauthor{#1}~\shortcite{#1}}

\def\Continue{\kw{continue}}

\begin{document}

\title{Cost Based Satisficing Search Considered Harmful}
\author{William Cushing \and J. Benton \and Subbarao
  Kambhampati\thanks{An
    extended abstract of this paper appeared in the proceedings of SOCS
    2010. This research is supported in part by ONR grants
    N00014-09-1- 0017 and N00014-07-1-1049, and the NSF grant
    IIS-0905672. }
  \\
  Dept. of Comp. Sci. and Eng. \\
  Arizona State University \\
  Tempe, AZ 85281 \\
}


\nocopyright

\maketitle

\begin{abstract}
  Recently, several researchers have found that cost-based satisficing
  search with A* often runs into problems. Although some "work
  arounds" have been proposed to ameliorate the problem, there has not
  been any concerted effort to pinpoint its origin.  In this paper, we
  argue that the origins can be traced back to the wide variance in
  action costs that is observed in most  planning domains.
  We show that such cost variance misleads A* search, and that this is
  no trifling detail or accidental phenomenon, but a systemic weakness
  of the very concept of ``cost-based evaluation functions +
  systematic search + combinatorial graphs''.  We show that
  satisficing search with sized-based evaluation functions is largely
  immune to this problem.

\end{abstract}

\section{Introduction}

Much of the scale-up, as well as the research focus, in the automated
planning community in the recent years has been on satisficing
planning. Unfortunately, there hasn't been a concomitant increase in
our understanding of satisficing search.  Too often, the ``theory'' of
satisficing search defaults to doing  A* with inadmissible heuristics. While
removing the requirement of admissible heuristics certainly relaxes
the guarantee of optimality, there is no implied guarantee of efficiency.  A
combinatorial search can be seen to consist of two parts: a
``discovery'' part where the (optimal) solution is found and a
``proof'' part where the optimality of the solution is verified. 
While an optimizing search depends crucially on both these phases, a
satisficing search is instead affected more directly by the discovery
phase. Now, standard A* search conflates the discovery and proof
phases together and terminates only when it picks the optimal path for
expansion.   By default, satisficing planners use the same search
regime, but relax the admissibility requirement on the heuristics. 
This may not cause too much of a problem in domains with
uniform action costs, but when actions can have non-uniform costs, the
the optimal and second optimal solution can be arbitrarily apart in
depth.  Consequently, A* search with cost-based evaluation functions 
can be an arbitrarily bad strategy for satisficing search, as it waits
until the  solution is both discovered {\em and} proved to be
optimal. 




To be more specific, consider a planning problem for which the 
cost-optimal and second-best solution to a problem exist on
10 and 1000 unspecified actions.  \emph{The optimal solution may be
  the larger one.}  How long should it take just to find the 10 action
plan?  How long should it take to prove (or disprove) its optimality?
In general (presuming PSPACE/EXPSPACE $\ne$ P):
\begin{enumerate}
\item Discovery should require time exponential in, at most, 10.
\item Proof should require time exponential in, at least, 1000.
\end{enumerate}
That is, in principle, the only way to (domain-independently) prove
that the 10 action plan is better or worse than the 1000 action one is
to in fact go and discover the 1000 action plan.  Thus, A* search with
cost-based evaluation function will take time proportional to $b^{1000}$
for either discovery or proof.

Using both abstract and benchmark problems, we will
demonstrate that this is a systematic weakness of any search that uses
cost-based evaluation function. In particular, we shall see that if
$\eps$ is the smallest cost action (after all costs are normalized so
the maximal cost action costs 1 unit), then the time taken to discover
a depth $d$ optimal solution will be $b^{d \over \eps}$. If all
actions have same cost, then $\eps \approx 1$ where as if the actions
have significant cost variance, then $\eps \ll 1$. We shall see that
for a variety of reasons, most real-world planning domains do exhibit
high cost variance, thus presenting an ``{\em $\eps$-cost trap}'' that
forces any cost-based satisficing search to dig its own
($\frac{1}{\eps}$ deep) grave.

Consequently, we argue that satisficing search should resist the
temptation to directly use cost-based evaluation functions (i.e., $f$
functions that return answers in cost units) even if they are
interested in the quality (cost measure) of the resulting plan.  We
will consider two size-based branch-and-bound alternatives: the
straightforward one which completely ignores costs and sticks to a
purely size-based evaluation function, and a more subtle one that uses
a cost-sensitive size-based evaluation function (specifically, the
heuristic returns the size of the cheapest cost path; see
Section~\ref{setup}). We show that both of these outperform
cost-based evaluation functions in the presence of $\eps$-cost traps,
with the second one providing better quality plans (for the same run
time limits) than the first in our empirical studies.



%






While some of the problems with cost-based satisficing search have
also been observed, in passing, by other researchers (e.g. \cite{benton10a,lama-journal}, 
and some work-arounds have
been suggested, our main contribution is to bring to the fore its
fundamental nature.  The rest of the paper is organized as follows. In
the next section, we present some preliminary notation to formally
specify cost-based, size-based as well as cost-sensitive size-based
search alternatives.  Next, we present two abstract and fundamental
search spaces, which demonstrate that \emph{cost-based} evaluation
functions are `always' needlessly prone to such traps
(Section~\ref{epsilon-cost-trap}).  Section~\ref{topological}
strengthens the intuitions behind this analysis by viewing A* search
as flooding topological surfaces set up by evaluation functions.  We
will argue that of all possible topological surfaces (i.e., evaluation
functions) to choose for search, cost-based is the worst.  In
Section~\ref{sec:empirical}, we put all this analysis to empirical
validation by experimenting with \acronym{LAMA} \cite{lama-journal}
and SapaReplan.  The experiments do show that size-based alternatives
out-perform cost-based search. Modern planners such as LAMA use a
plethora of improvements beyond vanilla A* search, and in the appendix
we provide a deeper analysis on which extensions of LAMA seem to help
it mask (but not fully overcome) the pernicious effects of cost-based
evaluation functions.

\mvp
\section{Setup and Notation}
\label{setup}


We gear the problem set up to be in line with the prevalent view of
state-space search in modern, state-of-the-art satisficing planners.
First, we assume the current popular approach of reducing planning to
graph search.  That is, planners typically model the state-space in a
causal direction, so the problem becomes one of extracting paths,
meaning plans do not need to be stored in each state.
More important is that the structure of the graph is given {\em
  implicitly} by a procedure $\Gamma$, the child generator, with
$\Gamma(v)$ returning the local subgraph leaving $v$; i.e.,
$\Gamma(v)$ computes the subgraph $(N^+[v],E(\aSet{v},V-v)) = (\aSet{
  u \mid vu \in E }+v, \aSet{ vu \mid vu \in E })$ along with all
associated labels, weights, and so forth. That is, our analysis
depends on the assumption that {\em an implicit representation of the
  graph is the only computationally feasible representation}, a common
requirement for analyzing the $A^*$ family of
algorithms~\cite{astar,dechter85}.



The search problem is to find a path from an initial state, $i$, to
some goal state in $\calG$.  Let costs be represented as edge weights,
say $c(uv)$ is the cost of an edge from $u$ to $v$.  Let $g_c^*(v)$ be
the (optimal) cost-to-reach $v$ (from $i$), and $h_c^*(v)$ be the
(optimal) cost-to-go from $v$ (to the goal).  Then $f^*_c(v) :=
g_c^*(v) + h_c^*(v)$, the cost-through $v$, is the cost of the
cheapest $i$-$\calG$ path passing through $v$.  For discussing
smallest solutions, let $f^*_s(v)$ denote the smallest $i$-$\calG$
path through $v$.  It is also interesting to consider the size of the
cheapest $i$-$\calG$ path passing through $v$, say $\hat{f}^*_{s}(v)$.

We define a search node $n$ as equivalent to a path represented as a
linked list. In particular, we distinguish this from the state of $n$
(its last vertex), $\attribii{n}{v}$. We say $\attribii{n}{a}$ (for
action) is the last edge of the path and $\attrib{n}{p}$ (for parent)
is the subpath excluding $\attribii{n}{a}$ and $\attrib{n}{v}$.  With
$\attribii{n}{a}$ an edge from $v$ to $u$ the function $g_c(n)$
($g$-cost) is just the recursive formulation of path cost: $g_c(n) :=
g_c(n.p) + c(vu)$ ($g_c(n) := 0$ if $n$ is the trivial path).  So
$g^*(v) \le g_c(n)$ for all $i$-$v$ paths $n$, with equality for at
least one of them. Similarly let $g_s(n) := g_s(n.p) + 1$ (initialized
at 0), so that $g_s$ is an upper bound on $g_s^*$.

A goal is a target vertex where a plan may stop and be a valid
solution. We fix a computed predicate $\calG(t)$ (a blackbox) encoding
the set of \emph{goal} vertices.  Let $h_c(v)$, the {\em heuristic},
be a procedure to estimate $h_c^*(v)$. We call $h_c$ \emph{admissible}
if it is a guaranteed lower bound. Let $h_s(v)$ estimate the remaining
depth to the nearest goal, and let $\hat{h}_{s}(v)$ estimate the
remaining depth to the cheapest reachable goal.

We focus on two different definitions of $f$ (the evaluation
function). Since we study cost-based planning, we consider $f_c(n) :=
g_c(n) + h_c(\attribii{n}{v})$; this is the (standard, cost-valued)
{\em evaluation function} of $A^*$: cheapest-completion-first.  We
compare this to $f_s(n) := g_s(n) + h_s(\attrib{n}{v})$, the canonical
size-valued (or search distance) evaluation function, equivalent to
$f_c$ under uniform weights. Any combination of $g_c$ and $h_c$ is
{\em cost-based}; any combination of $g_s$ and $h_s$ is {\em
  size-based} (e.g., breadth-first search is size-based).  The
evaluation function $\hat{f}_s(n) := g_s(n) +
\hat{h}_s(\attrib{n}{v})$ is also size-valued, but cost-sensitive and
preferable.

{\footnotesize
\begin{codebox}
 \Procname{$\proc{Best-First-Search}(i,\calG,\Gamma,h_c,\proc{evaluate})$}
\li  \proc{initialize-search}()
\li  \While \id{open} not empty \Do
\li      $n \gets \attribii{open}{\func{remove}}()$
\li      $s \gets \attribii{n}{v}$
\li      \If \proc{bound-test}() \Then \Continue \End \label{line:branch-and-bound}
\li      \If \proc{goal-test}() \Then \Continue \End 
\li      \If \proc{duplicate-test}() \Then \Continue \End 
\li      $\id{star} \gets \Gamma(s)$ \RComment{Expand $s$}
\li      \For each edge $a$ from $s$ to a child $s'$ in $\id{star}$ \Do
\li          $n' \gets n  \langle s  a  s' \rangle$ \RComment{Extend the path $n$}
\li          \fbox{$f \gets \proc{evaluate}(n')$}\rule[-8pt]{0pt}{20pt}
\li          $\attribii{open}{\func{add}}(n',f)$
         \End
     \End
\li  \Return $\id{best-known-plan}$ \RComment{Optimality is proven.}
\end{codebox}
}

\begin{codebox}
\Procname{$\proc{evaluate}(n)$}
\zi \Comment What is the best measure on paths, $\gamma()$, to use?
\li $s \gets \attribii{n}{v}$
\li $n' \gets \proc{relaxed-solve}(s,\calG,\dots)$
\li $f \gets \gamma(nn')$
\zi \Comment With $f = g + h$ the first variations to consider are:
\zi \Comment $g \gets g_c(n), \quad h \gets g_c(n'),$ and
\zi \Comment $g \gets g_s(n), \quad h \gets g_s(n').$
\li \Return f
\end{codebox}

\begin{codebox}
\Procname{\proc{initialize-search}()}
\li  \id{open} $\gets$ empty priority queue
\li  \id{closed} $\gets$ empty map from vertices to paths
\li  $f_c^+ \gets \infty$ \RComment{An upper bound on $f_c^*(i)$}
\li  $\id{best-known-plan} \gets \const{null}$
\li  $n \gets \langle i \rangle$
\li  \fbox{$f \gets \proc{evaluate}(n)$}\rule[-8pt]{0pt}{20pt}
\li  $\attribii{open}{\func{add}}(n,f)$
\end{codebox}

\begin{codebox}
\Procname{\proc{bound-test}()}
\zi  \Comment $h_c()$ {\bf must} be a lower bound on $h^*_c()$
\li  \Return $g_c(n) + h_c(s) \ge f_c^+$
\end{codebox}

\begin{codebox}
\Procname{\proc{goal-test}()}
\li      \If $\calG(s)$  \Then
\li          $f_c^+ \gets g_c(n)$
\li          $\id{best-known-plan} \gets n$
\li          report $\id{best-known-plan}$
\li          \Return \const{true} \End
\li      \Return \const{false}
\end{codebox}

\begin{codebox}
\Procname{\proc{duplicate-test}()}
\li      $n' \gets \attribii{closed}{\func{get}}(s)$
\li      \If $n'$ not null \Then
\li           \If $g_c(n') \le g_c(n)$ \Then 
\li               \Return \const{true} \End
\zi           \Comment Need to re-expand $s$, eventually.
\zi           \Comment Doing nothing here is one strategy.
         \End
\li      $\attribii{closed}{\func{put}}(s,n)$
\li      \Return \const{false}
\end{codebox}

Pseudo-code for best-first branch-and-bound search of implicit
graphs is shown above.  
It continues searching after a solution is encountered and uses the
current best solution value to prune the search space
(line~\ref{line:branch-and-bound}).  The search is performed on a
graph implicitly represented by $\Gamma$, with the assumption being
that the explicit graph is so large that it is better to invoke
expensive heuristics (\proc{evaluate}) during the search than it is to
just compute the graph up front. 
The pseudo-code given for \proc{evaluate} shows one particular
approach (solving relaxed problems) to automatically devising
guidance; in that setting, the question considered by this paper is
whether to measure the sizes or the costs of the two paths ($n$ and
$n'$).


With respect to normalizing costs, we can let $\eps:=\frac{\min_a
  c(a)}{\max_a c(a)}$, that is, $\eps$ is the least cost edge after
normalizing costs by the maximum cost (to bring costs into the range
$[0,1]$). 
We use the symbol $\eps$ for this ratio as we anticipate actions with
high cost variance in real world planning problems.  For example:
boarding versus flying (ZenoTravel), mode-switching versus machine
operation (Job-Shop), and (unskilled) labor versus (precious) material
cost. 




\mvp
\section{$\eps$-cost Trap: Two Canonical Cases}
\label{epsilon-cost-trap} 

In this section we argue that the mere presence of $\eps$-cost
misleads cost-based search, and that this is no trifling detail or
accidental phenomenon, but a systemic weakness of the very concept of
``cost-based evaluation functions + systematic search + combinatorial
graphs''. 
We base this analysis in two abstract search spaces, in order to
demonstrate the fundamental nature of such traps.  The first abstract
space we consider is the simplest non-trivial (non-uniform cost) search space, the search space of
a (large) cycle with one expensive edge.  The second abstract space we
consider is a more natural model of search (in planning), a uniform
branching tree.  Traps in these spaces are just exponentially sized
\emph{and connected} sets of $\eps$-cost edges: not the common result
of a typical random model of search (sampling edges independently). We briefly
consider why planning benchmarks naturally give rise to such structure.

\mvp
\subsection{Cycle Trap}

In this section we consider the simplest abstract example of the
$\eps$-cost `trap', where applying increasingly powerful heuristics and
domain analysis to ones search problem gives rise to an `effective
graph' --- the graph for which Dijkstra's algorithm produces
isomorphic behavior.  (In particular take $h=0$ in this section.)
Presumably such graphs have rather complex shape; but certainly
complex graphs contain simple graphs as subgraphs.  So if there is a
problem with search behavior in an exceedingly simple (non-uniformly
weighted) graph then we can suppose that no amount of domain analysis,
learning, heuristics, and so forth, will incidentally address the
problem: the inference must specifically address the issue of
non-uniform weights.  So we are arguing that $\eps$-cost is by itself
a fundamental challenge to be overcome in planning: unsubsumed by
other challenges.

The state-space we will consider is the cycle, with an associated
exceedingly simple metric consisting of all uniform weights but for a
single expensive edge.
There are several other candidates for simple non-trivial state-spaces
(e.g., cliques), but clearly the cycle is fundamental.  Its search
space is certainly the simplest non-trivial search space: the rooted
tree on two leaves.  So the single decision to be made is in which
direction to traverse the cycle: clockwise or counter-clockwise.
Formally:

\noindent{\bf $\varepsilon$-cost Trap:} Consider the problem of making
some counter, say $x$, on $k$ bits contain one less than its maximum
value ($2^k-2$), starting from $0$, using only the operations of
increment and decrement.  There are 2 minimal solutions: incrementing
$2^k-2$ times, or decrementing twice (exploiting overflow).  Set the
cost of incrementing and decrementing to 1, except that overflow (in
either direction) costs, say, $2^{k-1}$.  Then the 2 minimal solutions
cost $2^k-2$ and $2^{k-1}+1$, or, normalized, $2(1-\varepsilon)$ and
$1+\varepsilon$.

Cost-based search is the clear loser on this problem.  While both
approaches prove optimality in exponential time ($O(2^k)$), size-based
discovers the optimal plan in constant time. Of course the goal
$2^k-2$ is chosen to best illustrate the trap.  So consider the
discovery problem for other goals: from $2^k [0,\frac{1}{2}]$
cost-based search is twice as fast, from $2^k
[\frac{1}{2},\frac{2}{3}]$ the performance gap narrows to break-even,
and from $2^k[\frac{2}{3},1)$ the size-based approach takes the lead
--- \emph{by an enormous margin}.  Note that between
$2^k[\frac{2}{3},\frac{3}{4}]$ there is a trade-off: size-based finds a
solution before cost-based, but cost-based finds the optimal solution
first.  (Of course, time till optimality is proven monotonically
favors the cost-based approach: by a factor of 2 in the region
$2^k[0,\frac{1}{2}]$, by a factor of 1 in the region
$2^k[\frac{3}{4},1)$, and by $1<(\frac{1}{2}+2\alpha)^{-1}<2$ for
goals of the form $2^k(\frac{1}{2}+\alpha)$.)

Then, even across all goals, cost-based search is still quite
inferior: the margins of victory either way are extremely lopsided.
To illustrate, consider `large' $k$, say, $k=1000$.  Even the most
patient reader will have forcibly terminated either search \emph{long}
before receiving any useful output --- except if the goal is of the
form $0 \pm f(k)$ for some sub-exponential $f(k)$.  Both approaches
discover and prove the optimal solution in the positive case in time
$O(f(k))$ (with size-based performing twice as much work).  In the
negative case, only the size-based approach manages to discover a
solution (in time $O(f(k))$) before being killed. Moreover, while it
will fail to produce a proof before death, we, based on superior
understanding of the domain, can show it to be \emph{posthumously}
correct (and have: $2^k-f(k) > 2^k\,\frac{3}{4}$ for large $k$).

In summary, cost-based search on the single-decision tree ``only
explores left''.  Hence the trap: There is no reason to suppose
that one direction is much worse than another in very large,
weighted, graphs just because the first step is quite expensive.

\mvp
\subsection{Branching Trap}

In the counter problem the trap is not even \emph{combinatorial}; the
search problem consists of a single decision at the root, and the trap
is just an exponentially deep path.  Then it is abundantly clear that
appending Towers of Hanoi to a planning benchmark, setting its
actions at $\eps$-cost, will kill cost-based search --- even given
the perfect heuristic for the puzzle!  \Ignore{(Humans, too, can be
  susceptible to distraction-by-puzzle, but not quite so
  dramatically.)}  Besides Hanoi, though, exponentially deep paths are
not typical of planning benchmarks.  So in this section we demonstrate
that exponentially large subtrees on $\eps$-cost edges are also traps.

Consider $x>1$ high cost actions and $y>1$ low cost actions in a uniform branching tree
model of search space.  (A typical model for analysis, appropriate up
to the point where duplicate state checking becomes significant.
See~\cite{pearl-heuristics} for similar analysis on more complex models of search.)
Suppose the solution of interest costs $C$, in normalized 
units, so the solution lies at depth $C$ or greater.  Then cost-based
search faces a grave situation: $O((x+y^{\frac{1}{\eps}})^C)$
possibilities will be explored before considering all potential
solutions of cost $C$.

A size-based search only ever considers
at most $O((x+y)^d)=O(b^d)$ possibilities before consideration of
all potential solutions of size $d$; of course the more
interesting question is how long it takes to find solutions of fixed
cost rather than fixed depth---note $\frac{C}{\eps} \ge d \ge C$.  
Assuming the high cost actions are relevant, that is, some number of them
are needed by solutions, then we have that solutions are not
actually hidden as deep as $\frac{C}{\eps}$.  Suppose, for
example, that solutions tend to be a mix of high and low cost actions
in equal proportion.  Then the depth of those solutions with cost $C$
is $d=2\,\frac{C}{1+\eps}$ ($\frac{d}{2} \cdot 1 + \frac{d}{2}
\cdot \eps = C$).  At
such depths the size-based approach is the clear winner: $O((x+y)^{\frac{2C}{1+\eps}}) \ll
O((x+y^{\frac{1}{\eps}})^C)$ (normally).  Consider, say, $y=\frac{b}{2}$, then:
\begin{align*}
b^{\frac{2C}{1+\eps}} / \left(x + y^{\frac{1}{\eps}}\right)^C 
&< b^{\frac{2C}{1+\eps}} / y^{\frac{C}{\eps}},\\
&< 2^{\frac{C}{\eps}} / b^{\frac{C}{\eps}\frac{1-\eps}{1+\eps}},\\
&< \frac{2}{b^{\frac{1-\eps}{1+\eps}}}^{\frac{C}{\eps}},
\end{align*}
and, provided $\eps < \frac{1-\log_b 2}{1+\log_b 2}$ (for $b=4$,
$\eps < \frac{1}{3}$), the last is always less than 1 and, for that
matter, goes, quickly, to 0 as $C$ increases and/or $b$ increases and/or $\eps$ decreases.

Generalizing, the size-based approach is faster at finding solutions of any given cost,
as long as (1) high-cost actions constitute at least some constant fraction of the solutions
considered, (2) the ratio between high-cost and low-cost is
sufficiently large,(3) the effective search graph (post additional
inference) is reasonably well modeled by an infinite uniform branching
tree (i.e., huge), and (4) the search is systematic.



%
%
%
\mvp
\section{Search Effort as Flooding Topological Surfaces of Evaluation Functions}
\label{topological}

We view evaluation functions ($f$) as topological surfaces over
search nodes, so that generated nodes are visited in, roughly, order of
$f$-altitude.  With non-monotone evaluation
functions, the set of nodes visited before a given node is all those
contained within some basin of the appropriate depth --- picture water
flowing from the initial state: if there are dams then such a
flood could temporarily visit high altitude nodes before low altitude
nodes.  (With very inconsistent heuristics --- large heuristic weights
--- the metaphor loses explanatory power, as there is nowhere to go
but downhill.  See~\cite{dechter85} for comprehensive details.)  

If we take a single point inside such a basin (but not one defining
the brim) and alter its altitude over the entire range of that basin's
depth, we will not have changed the set of states inundated prior to
the brim.  If there were no solutions prior to the brim, then we will
not have altered any externally visible behavior of the search:
Whenever best-first search finally finds a solution it will no longer
have mattered how all the prior nodes were ordered.  To illustrate,
$\text{IDA}^*$ deserves its name, despite exploring the space in an entirely
different order from $A^*$ in any given iteration.  

In particular, controlling the behavior of search by altering the
evaluation function is a very different proposition in the two
contexts of local search and best-first search.  For the latter,
preventing exploration of some choice requires raising its altitude
(or that of a cut-set) to past that of a solution of interest,
actually, past the altitude of every cut-set separating that solution from the initial vertex (the altitude of a
set is the minimum over its elements), i.e., past the rim of the
deepest basin preventing inundation of the solution.  For the
former, mitigating exploration is merely a matter of making the choice
worse than its best sibling; the ideal amount of penalization
depends on the nature of randomization applied.\footnote{The second
best sibling could be second most likely to be chosen, but it could also be the
least likely to be chosen.}

Formally, but with narrower context: Consider an $h_c$ that is derived
by optimally solving relaxed problems, or just directly suppose that
$h_c$ is guaranteed to be admissible and
consistent~\cite{pearl-heuristics}.  Consider the altitude
($f_c^*(i)$) of the cost-optimal solution in $f_c$.  All
lower-altitude nodes comprise the \emph{cost-optimal footprint}.
Exhausting the footprint is a proof, relative to $h_c$ being
admissible, of the purported optimality of the known solution (with
$h_c$ consistent, exhaustion is moreover necessary for proof by
search).  As the order of doing so does not affect correctness of the
proof, there is significant freedom/futility (depending on your
perspective) in the choice of evaluation function: Every systematic
search is equivalent (does the same amount of total work) if $h_c$ and
$f_c^*(i)$ are given. When re-expansion is a significant possibility,
then the appropriate statement is that the same set of states are
expanded, some, hopefully few, more than once.  It follows that
performing two levels of search, the outer search taking guesses at
$f_c^*(i)$, is a powerful idea (as in $IDA^*$, or in the standard
treatment of optimization problems as decision problems).

That is, it is futile to attempt to expand less than $A^*$, but, one
is free to expand that set in any order.  For example, with an
oracular guess of $f_c^*(i)$, it is possible to terminate in equal time
yet print the optimal solution sooner than $A^*$:  take the evaluation
function to be $-f_c$, so that the optimal solution is expanded as
soon as it is generated, at which time, perhaps, the open list still
contains some states with $-f_c(s) > -f_c^*(i)$.  Indeed, as the optimal
solution is guaranteed to be the last path expanded, up to tie-breaking,
under the evaluation function $f_c$, \emph{any} (other) evaluation function
(monotonically) improves upon the performance of $A^*$, with respect
to the problem of finding the optimal solution.

\noindent {\bf Worst-case:} The minimum gradient in $g$ bounds the
worst-case of the discovery problem: it puts a limit on the number of
search nodes that could conceivably be considered just as good as some
solution of interest.  For example, in uniformly branching trees the
absolute worst-case bound is $b^{d\frac{\max\grad g}{\min\grad g}}$
(with $d$ the depth of the unique solution).  Insisting on a fairer
distribution of edge costs and/or considering non-zero heuristics (but
still imperfect) lowers the bound, but not asymptotically: still
$O(b^{d\frac{\max\grad g}{\min\grad g}})$ many search nodes might be
expanded before finding the solution (in the worst-case of a unique
solution on $d$ maximum cost actions).  Other search models yield
different bounding expressions, but all will be increasing functions
of $d\frac{\max \grad g}{\min \grad g}$.  Considering normalized
representations then $\max \grad g$ is just $1$, and so we have that
$f_s$ enjoys the tightest bound, since $\min \grad g_s = 1$.  In
contrast, $f_c$ suffers from the `loosest' bound, as $\min \grad g_c =
\eps \ll 1$, in the sense that one presumably devotes bits to
specifying costs (in binary), so one cannot do worse than
exponentially small except by permitting zero costs.  Taking
worst-case for some specific $f$ to mean a problem with maximum search
nodes at every altitude, with a unique solution of maximum cost (given
its size), then, for the discovery problem: (1) Size-based search
achieves the asymptotically best-possible worst-case performance. (2)
Cost-based search `achieves' the asymptotically \emph{worst}-possible
worst-case performance.

Note that all that is being said is that a malicious problem-setter
has control of the metric, so any quality-sensitive search can be
misdirected.  

\noindent {\bf Typical-case:}
Every choice of search topology will eventually lead to identification
of the optimal solution and exhaustion of the cost-optimal footprint.
Some will produce a whole slew of suboptimal solutions along the way,
eventually reaching a point where one begins to wonder if the most
recently reported solution is optimal.  Others report nothing until finishing.  The former are
interruptible, and are rather more desirable than the latter.
That is, admissible cost-based topology is the worst possible choice:
it is the \emph{least} interruptible. There is no point at which one
can forcibly terminate and receive anything\footnote{Besides a better
  lower bound.} for ones investment of
computational resources.  Gaining interruptibility is a matter of raising the altitude of large
portions of the footprint in exchange for lowering the altitude of a
smaller set of non-footprint search nodes (leaving the solution of
interest fixed).  Note that there must be a trade-off (else one has
devised a better heuristic): interruptibility comes at the expense of
total work.  

With size-based topology, the large set is the set of \emph{longer}
yet \emph{cheaper} plans, while the small set is the \emph{shorter}
yet \emph{costlier} plans.  In general one expects there to be many
more longer plans than shorter plans in combinatorial problems, but
that changes if the problem is hardest possible in finite spaces,
i.e., all goal states are as far away as possible so that cheap
solutions are also necessarily long.  There is no reason to suppose
that size-based topology is the best possible trade-off; it just
demonstrates existence of better approaches than admissible cost-based
topology.  Inadmissible cost-based topology, such as $WA^*$, can also
demonstrate existence of better approaches.

Weighting the heuristic, though, magnifies depth-first behavior, which is great
up until finding a solution, but afterwards leads to poor backtracking
behavior.  For example, depth-first bias in a non-uniformly weighted
uniform branching tree permits catastrophic backtracking
behavior: exhaustion of maximum size $\eps$-cost traps. (And tree models are better fits under depth-first bias, as
state re-expansion is more likely due to finding better paths later.)  Dynamically weighting the heuristic is one approach~\cite{dyn-wastar}, attacking the
contribution that non-uniform accuracy of heuristics has on such
backtracking, one could also consider randomized restarts of $WA^*$
along with a decreasing schedule of weights.\footnote{The possibility of
  state re-expansion greatly exacerbates poor backtracking behavior,
  so it is worthwhile to keep in mind that an iterated search need not
  re-expand states immediately.}  Employing multiple open lists (as in
\acronym{LAMA}) is a different approach (than restarting) to
permitting non-local backtracking; \acronym{EES}~\cite{thayer-ees}
does so while also, unlike the preceding, explicitly considering the further
impact that non-uniform weights have, achieving an interesting blend
of cost and size considerations.  One could characterize it as
cost-bounded size optimization; it is also interesting to
consider reformulating \acronym{EES} as size-bounded cost
optimization, particularly considering the behavior of
GraphPlan/BlackBox~\cite{graphplan,blackbox} and relatives.

\mvp
\section{$\eps$-cost Trap in Practice}
\label{sec:empirical}

\begin{figure}[t]
\begin{center}
\includegraphics[clip=true,trim=0in 0.6in 0in 0in,width=2.7in,height=1.8in]{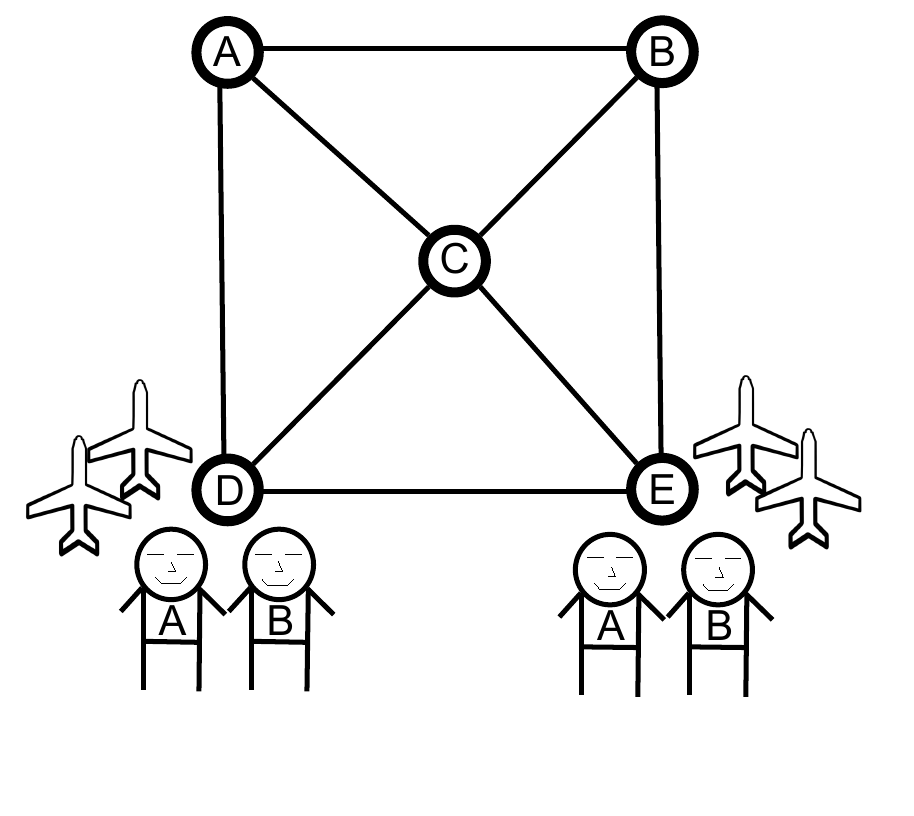}
\end{center}
\caption{\label{fig:travel}Rendezvous problems. Diagonal edges cost
  7,000, exterior edges cost 10,000. Board/Debark cost 1.\vspace{-0.15in}}
\end{figure}

In this section we demonstrate existence of the problematic planner
behavior in a realistic setting: running LAMA on problems in the
travel domain (simplified ZenoTravel, zoom and fuel removed), as well
as two other IPC domains. 
Analysis of LAMA is complicated by many factors, so we also test the
behavior of SapaReplan on simpler instances (but in all of
ZenoTravel).  The first set of problems concern a rendezvous at the
center city in the location graph depicted in Figure~\ref{fig:travel};
the optimal plan arranges a rendezvous at the center city.  The second
set of problems is to swap the positions of passengers located at the
endpoints of a chain of cities.

\mvp
\subsection{LAMA}

In this section we demonstrate the performance
problem wrought by $\eps$-cost in a state-of-the-art (2008)
planner --- LAMA~\cite{lama-journal}, the leader of the cost-sensitive
(satisficing) track of IPC'08~\cite{ipc08}.
With a completely trivial recompilation (set a flag) one can
make it ignore the given cost function, effectively searching by
$f_s$.  With slightly more work one can do better
and have it use $\hat{f}_s$ as its evaluation function, i.e., have the
heuristic estimate $\hat{d}$ and the search be size-based, but still
compute costs correctly for branch-and-bound.  Call this latter modification LAMA-size.
Ultimately, the observation is that LAMA-size outperforms LAMA --- no
trivial feat, particularly for such a small change in implementation.

LAMA\footnote{Options: `fFlLi'.} defies analysis in a number
of ways: \emph{landmarks},\Ignore{\footnote{``Landmark analysis'' is a misnomer: in standard usage a landmark is a \emph{sufficient}
condition  (for either dead-ends, ``\dots then you've gone too far'',
or achievement, ``\dots then you can't miss it''), but in automated planning the
jargon stands only for a \emph{necessary} condition.  Principled use
of these, as LAMA does, is then not an \emph{especially} promising attack upon search plateaus.}} \emph{preferred operators}, \emph{dynamic evaluation
  functions}, \emph{multiple open lists}, and \emph{delayed
  evaluation}, all of which effect potential search plateaus in complex ways.
Nonetheless, it is essentially a cost-based approach.

\Ignore{
For the first several stages the
search space is that induced by only \emph{preferred operators},
afterwords, all operators (actions) are considered.  Eventually the heuristic
weight decreases to 1, and remains constant in all later stages; there
are potentially multiple such stages as the heuristic(s) are not
(assumed to be) admissible.\footnote{Restarting is one approach to
  branch-and-bound; restarting is particularly appropriate if the evaluation
  function changes upon finding solutions.   Once the heuristic weight
  is constant, though, it may no longer make sense to restart.}
LAMA uses \emph{multiple open lists} (two, with differing heuristics), with search nodes being
entered into every open list, avoiding needless re-expansion by exploiting the closed list.  (So duplicate
states may induce re-expansion, if the cheaper path is found second,
but not duplicate search nodes.)  LAMA also employs \emph{delayed evaluation}, meaning
that the $f$-value actually assigned is based on the $g$-value of the
node in question and the $h$-value of its parent (so the search
assumes no child makes progress w.r.t.\ $h$).  As a result, siblings are visited
cheapest first.}

\noindent{\bf Results.}\footnote{New best plans
  for Elevators were found (largely by LAMA-size).  The baseline planner's score is 71.8\% against the better reference plans.}  With more than about 8 total passengers, LAMA
is unable to complete any search stage except the first (the greedy search). 
For the same problems, LAMA-size finds the same first plan (the
heuristic values differ, but not the structure), but is then subsequently able to
complete further stages of search.  In so doing it sees marked
improvement in cost; on the larger problems this is due only to
finding better variants on the greedy plan.  Other domains are
included for broader perspective, woodworking in particular was chosen
as a likely counter-example, as all the actions concern just one
type of physical object and the costs are not wildly different.  For
the same reasons we would expect LAMA to out-perform LAMA-size in some
cost-enhanced version of Blocksworld.  For a comprehensive empirical
analysis, see~\cite{lama-journal}.

\begin{table}[t]
\centerline{\begin{tabular}{|c||c|c|}
\hline
Domain & LAMA & LAMA-size\\
\hline
Rendezvous              & 70.8\% &83.0\%\\
Elevators               & 79.2\% &93.6\%\\
Woodworking             & 76.6\% &64.1\%\\
\hline
\end{tabular}}
\caption{\label{tbl:lama}IPC metric on LAMA variants.}
\vspace{-0.15in}
\end{table}

\mvp
\subsection{SapaReplan}

We also consider the behavior of SapaReplan on the simpler set of problems.\footnote{Except that these
  problems are run on all of ZenoTravel.}  This planner is much less
sophisticated in terms of its search than LAMA, in the sense of being much closer to a
straight up implementation of weighted A* search.  The problem is just to swap the locations of passengers located on either side of a
chain of cities.  A plane starts on each side, but there is no actual
advantage to using more than one (for optimizing either of size or
cost): the second plane exists to confuse the planner.
Observe that smallest and cheapest plans are the same.  So in some
sense the concepts have become only superficially different; but this
is just what makes the problem interesting, as despite this
similarity, still the behavior of search is strongly affected by the nature of
the evaluation function.  We test the performance of $\hat{f}_s$ and $f_c$,
as well as a hybrid evaluation function similar to $\hat{f}_s + f_c$ (with costs
normalized).  We also test hybridizing via tie-breaking conditions,
which ought to have little effect given the rest of the search framework.

\noindent{\bf Results.}\footnote{The results differ markedly between the 2 and 3 city sets of problems
because the sub-optimal relaxed plan extraction in the 2-cities problems
coincidentally produces an essentially perfect heuristic in many of
them.  One should infer that the solutions found in the 2-cities
problems are sharply bimodal in quality and that the meaning of the average is
then significantly different than in the 3-cities problems.} The
size-based evaluation functions find better cost plans faster (within
the deadline) than cost-based evaluation functions.  The hybrid
evaluation function also does relatively well, but not as well as
could be hoped. Tie-breaking has little effect, sometimes negative.

We note that Richter and Westphal (2010) also report
that replacing cost-based evaluation function with a pure size-based one
improves performance over LAMA in multiple other domains. Our version
of LAMA-Size uses a cost-sensitive size-based search, and our results,
in the domains we investigated, seem to show bigger improvements over
LAMA.

Finally, while LAMA-size outperforms LAMA, our theory of $\eps$-cost
traps suggests that cost-based search should fail even more
spectacularly. In the appendix, we take a much closer look at one
domain--the travel domain--and present a detailed study of which
extensions of LAMA help it temporarily mask the pernicious effects of
cost-based search. Our conclusion is that both LAMA and SapaReplan
manage to find solutions to problems in the travel domain despite the
use of a cost-based evaluation function by using various tricks to
induce a limited amount of {\em depth-first behavior} in an
$A^*$-framework.  This has the potential effect of delaying
exploration of the $\eps$-cost plateaus slightly, past the discovery
of a solution, but still each planner is ultimately trapped by such
plateaus before being able to find really good solutions.  In other
words, such tricks are mostly serving to mask the problems of
cost-based search (and $\eps$-cost), as they merely delay failure by
just enough that one can imagine that the planner is now effective
(because it returns a solution where before it returned none).  Using
a size-based evaluation function more directly addresses the existence
of cost plateaus, and not surprisingly leads to improvement over the
equivalent cost-based approach --- even with LAMA.

\begin{table}[t]
\centerline{\begin{tabular}{|c||c|c|c|c|}
\hline
 & \multicolumn{2}{c|}{2 Cities} & \multicolumn{2}{c|}{3  Cities}\\
\hline
Mode & Score & Rank & Score & Rank\\
\hline
Hybrid                  & 88.8\% & 1& 43.1\% &2\\
Size                    & 83.4\% & 2& 43.7\% &1\\
Size, tie-break on cost & 82.1\% & 3& 43.1\% &2\\
Cost, tie-break on size & 77.8\% & 4& 33.3\% &3\\
Cost                    & 77.8\% & 4& 33.3\% &3\\
\hline
\end{tabular}}
\caption{IPC metric on SapaReplan variants in
  ZenoTravel.}
\vspace{-0.15in}
\end{table}

%
%

\mvp
\section{Conclusion}

The practice of combinatorial search in automated planning is
\emph{satisficing}.
There is a great call for deeper theories of satisficing search, and
one perhaps significant obstacle in the way of such research is the
pervasive notion that perfect problem solvers are the ones giving only
perfect solutions.  Actually implementing cost-based, systematic,
combinatorial, search reinforces this notion, and therein lies its
greatest harm.  

In support of the position we demonstrated the technical difficulties arising from such use of
a cost-based evaluation function, largely by arguing that the
size-based alternative is a notably more effective default strategy.  We
argued that using cost as the basis for plan evaluation is a purely
exploitative perspective, leading to least interruptible behavior.  Being least
interruptible, it follows that implementing cost-based search will
typically be immediately harmful to that particular application.  But
regardless of whether the particular instance demonstrates the rule or
the exception, the lasting harm is in reinforcing the wrong definition
of \emph{satisficing search} in the first place.  In conclusion, as a
rule: Cost-based search is harmful.

\bibliographystyle{aaai}
{\footnotesize \bibliography{fsize}}

\appendix

\section{Deeper Analysis of the Results in Travel Domain}

In this section we analyze the reported  behavior of LAMA and SapaReplan in
greater depth.  We begin with
a general analysis of the domain itself and the behavior of
(simplistic) systematic state-space search upon it, concluding that cost-based
methods suffer an enormous disadvantage.  The empirical results are
not nearly so dramatic as the dire predictions of the theory, or at
least do not appear so.  We consider to what extent the various
additional techniques of the planners (violating the assumptions of
the theory) in fact mitigate the pitfalls of $\eps$-cost, and to what
extent these only serve to mask the difficulty.

\subsection{Analysis of Travel Domain}

We argue that search under $f_c$ pays a steep price in time and memory
relative to search under $\hat{f}_s$.  The crux of the matter is that
the domain is reversible, so relaxation-based heuristics cannot
penalize fruitless or even counter-productive passenger movements by
more than the edge-weight of that movement.\Ignore{\footnote{It
    achieves little to forbid immediate reversal in the heuristic, and
    it is very difficult to domain-independently justify completely
    leaving the inverse actions out.}} Then plateaus in $g$ are
plateaus in $f$, and the plateaus in $g_c$ are enormous.

First note that the domain has a convenient structure: The global state
space is the product of the state space of shuffling planes around
between cities/airports via the fly action (expensive), and the state space of shuffling
people around between (stationary) planes and cities/airports via the
board/debark actions (cheap).  For example, in the rendezvous problems, there are $5^4=625$ possible assignments
of planes to cities, and $(5+4)^{2k}$ possible assignments of
passengers to locations (planes + cities), so that the global state
space has exactly $5^4 \cdot 9^{2k}$ reachable states (with $k$ the number of
passengers at one of the origins).\footnote{Fuel and zoom are
  distracting aspects of ZenoTravel-STRIPS, so we remove
  them.  Clever domain analysis could do the same.}\Ignore{For $k=6$ (12 passengers, 4 planes, and 5 cities) this is already well over a
trillion states. Leaving planes stationary, only $2^{2k}$ passenger assignments are actually reachable
from a `typical' plane assignment; this is the subspace of
passenger-shuffling from such a plane assignment.  (The initial state is `atypical'
with $3^{2k}$ reachable states in its passenger-shuffling subspace.)  }

Boarding and debarking passengers is extremely cheap, say on the order
of cents, while flying planes between cities is quite a bit more
expensive, say on the order of hundreds of dollars (from the
perspective of passengers).\Ignore{\footnote{So a more accurate model
    includes tickets, or takes flying cost as a multiple of passengers
    onboard.}}  So $\frac{1}{\eps}\approx 10000$ for this domain --- a
constant, but much too large to ignore.

To analyze state-space approaches in greater depth let us make all of the following additional assumptions: 
The heuristic is relaxation-based, imperfect, and in particular heuristic error
is due to the omission of actions from relaxed solutions relative to
real solutions.  Heuristic error is not biased in favor of less error in
  estimation of needed fly actions --- in this problem planes are
  \emph{mobiles} and \emph{containers} whereas people are only
  \emph{mobiles}.  Finally, there are significantly but not overwhelmingly more passengers
  than planes.

Then consider a child node, in plane-space, that is in fact the
correct continuation of its parent, but the heuristic fails to realize
it.  So its $f$ is higher by the cost or size of one plane movement:
$1$ under normalized costs.  
Moreover assume that moving passengers is not heuristically good
(in this particular subspace).  (Indeed, moving passengers is usually
a bad idea.)  Then moving a passenger increases $f_c$ by at most $2\eps$ (and at
least $\eps$), once for $g_c$
and once for $h_c$.  As $\frac{1}{2\eps}\approx 5000$ we have that
search under $f_c$ explores the passenger-shuffling space of the
parent to, at least, \emph{depth} 5000. Should the total heuristic error in
fact exceed one fly action, then each such omission will induce
backtracking to a further 5000 levels: for any search node $n$ reached
by a fly action set $e_c(n) = f_c(x) - f_c(n)$ with $x$ some
solution of interest (set $e_s$ similarly).  Then if
search node $n$ ever appears on the open list it will have its passenger-shuffling
subspace explored, under $f_c$, to at least depth $e_c \cdot 5000$
before $x$ is found (and at most depth $e_c \cdot \frac{1}{\eps}$).
Under $\hat{f}_s$, we have instead exploration up to at least depth $e_s
\cdot \frac{1}{2}$ and at most depth $e_s \cdot \frac{1}{1}$.

As 5000 objects is already far above the
capabilities\Ignore{\footnote{For that matter, the domain model is far from
  correct for thousands of passengers: capacities of planes are not
  modeled.}}\Ignore{  So one can use just a single plane, and indeed, 
  automated satisficing planners have a penchant for doing so.
  Considering such a model, the situation is only worse for cost-based
search as its depth of exploration will no longer be independent of
heuristic error: instead it really will explore thousands of levels deep.} of any current domain-independent planners, we can
say that at most plane-shuffling states considered, cost-based search
\emph{exhausts} the entire associated passenger-shuffling space during
backtracking.  That is, it stops exploring the space due to exhausting
finite possibilities, rather than by adding up sufficiently many
instances of $2\eps$ increases in $f$ --- the result is the same as if
the cost of passenger movement was 0.  Worse, such exhaustion
commences immediately upon backtracking for the first time (with
admissible heuristics).  Unless \emph{very} inadmissible (large
heuristic weights), then even with inadmissible heuristics, still
systematic search should easily get trapped on cost plateaus ---
before finding a solution.

In contrast, size-based search will be exhausting only those passenger
assignments differing in at most $e_s$ values; in the \emph{worst} case this
is equivalent to the cost-based method, but for good heuristics is a
notable improvement.  (In addition the size-based
search will be exploring the plane-shuffling space deeper, but that space is
[assumed to be] much smaller than any single passenger-shuffling
space.)  Then it is likely the case that cost-based search dies before
reporting a solution while size-based search manages to find one or more.

\subsection{Analyzing LAMA's Performance}
While LAMA-size out-performs LAMA, it is hardly as dramatic a difference as predicted above.  Here
we analyze the results in greater depth, in an attempt to understand
how LAMA avoids being immediately trapped by the passenger-shuffling
spaces.  Our best, but not intuitive, explanation is its
pessimistic delayed evaluation leads to a temporary sort of
depth-first bias, allowing it to skip exhaustion of many of the
passenger-shuffling spaces until \emph{after} finding a solution.  So,
(quite) roughly, LAMA is able to find one solution, but not two.  

\noindent{\bf Landmarks.} The passenger-shuffling subspaces are search
plateaus, so, the most immediate hypothesis is
that LAMA's use of landmarks helps it realize the futility of large
portions of such plateaus (i.e., by pruning them).  However, LAMA uses landmarks only as a
heuristic, and in particular uses them to order an additional (also
cost-based) open list (taking every other expansion from that list), and the end result is actually greater breadth of
exploration, not greater pruning.  

\noindent{\bf Multiple Open Lists.} Then an alternative hypothesis is
that LAMA avoids immediate death by virtue of this additional
exploration, i.e., one open list may be stuck on an enormous search
plateau, but if the other still has guidance then potentially LAMA can
find solutions due to the secondary list.  In fact, the lists interact
in a complex way so that conceivably the multiple-list approach even
allows LAMA to `tunnel' out of search plateaus (in either list, so
long as the search plateaus do not coincide).  Indeed the secondary list improves performance, but turning it off
still does not cripple LAMA, let alone outright kill it.

\Ignore{First let us note that the greedy search is remarkably consistent in
its behavior across variations on the theme of the reported problem:
if the greedy solution moves some plane to both destinations, then it
does so in the wrong order.  The wrong
order is to fly across the diagonal and then backtrack along the top
edge; the correct order, saving 20\%, is to fly around 2 sides of the
square.  It seems that the reason is only that the up-front cost of
flying to the center is cheaper.  However, it is unclear whether that
is due to the search preferring low-cost or the heuristics preferring
low-cost --- at that point, preferred operators are on, so it is
difficult to characterize even just the search-space itself.
Turning preferred operators off, though, still exhibits this strong
preference for flying along the diagonal edges (and for boarding
most/all passengers before any flying) --- even for such
subgoals as moving along a single side of the square, the plans
produced fly to the center first, instead.  (That would be a
potentially good decision if passenger movements were coordinated
across planes, but that is not what is taking place in the greedy plans.)
}

\noindent{\bf Small Instances.} It is illuminating to consider the behavior of LAMA and LAMA-size with only 4
passengers total; here the problem is small enough that optimality can be proved.  LAMA-size
terminates in about 12 minutes.  LAMA terminates in about 14.5
minutes.  Of course the vast majority of time is spent in the last
iteration (with heuristic weight 1 and all actions considered) --- and both are unrolling the exact same portion of state
space (which is partially verifiable by
noting that it reports the same number of unique states in both
modes).  There is only one way that such a result is at all possible:
the cost-based search is re-expanding many more states.  That is difficult to believe; if anything it is the size-based
approach that should be finding a greater number of suboptimal
paths before hitting upon the cheapest.  The explanation is two-fold.  First of all pessimistic
delayed evaluation leads to a curious sort of depth-first behavior.  Secondly, cost-based search pays far more
dearly for failing to find the cheapest path first.

\noindent{\bf Delayed Evaluation.}  LAMA's delayed
evaluation is not equivalent to just pushing the original search
evaluation function down one level.  This is because it is the
\emph{heuristic} which is delayed, not the full evaluation function.  LAMA's evaluation function is the sum of the
\emph{parent's} heuristic on cost-to-go and the \emph{child's}
cost-to-reach: $f_L(n) = g(n) + h(n.p.v)$.  One can view this technique, then, as a transformation
of the original heuristic.  Crucially, the technique increases the
inconsistency of the heuristic.  Consider an optimal path and the
perfect heuristic.  Under delayed evaluation of the perfect heuristic, each
sub-path has an $f_L$-value in excess of $f^*$ by exactly the cost of
the last edge.  So a high cost edge followed by a low cost edge
demonstrates the non-monotonicity of $f_L$ induced by the inconsistency
wrought by delayed evaluation.  The problem with non-monotonic
evaluation functions is not the decreases \emph{per se}, but the increases that precede them.  In this case,
a low cost edge followed by a high cost edge along an optimal path
\emph{induces backtracking} despite the perfection of the heuristic
prior to being delayed.

\Ignore{Consider a pessimistic delayed evaluation applied to $h^*$.  So the
eager evaluation function is perfect.  Along an optimal path, the new
$f$-values are the old $f$-values (all tied at $f^*$) plus the cost of the last edge in
the subpath.  Then low-cost edges following high-cost edges produce
marked inconsistency in the evaluation function (evaluation functions
are consistent iff monotone).  Still inconsistency only makes it
possible, not guaranteed, to find sub-optimal paths first.  }

\noindent{\bf Depth-first Bias.} 
Consider some parent $n$ and two children $x$ and $y$ ($x.p = n$, $y.p
= n$) with $x$ reached by some cheap action and $y$ reached by some
expensive action.  Observe that siblings are always expanded in order of
their cost-to-reach (as they share the same heuristic value), so $x$
is expanded before $y$.  Now, delaying evaluation of the heuristic was
pessimistic: $h(x.v)$ was taken to be $h(n.v)$, so that it appears
that $x$ makes no progress relative to $n$.  Suppose the pessimism was
unwarranted, for argument's sake, say entirely unwarranted: $h(x.v) =
h(n.v) - c(x.o)$.  Then consider a cheap child of $x$, say $w$.  We
have:
\begin{align}
f_L(w) &= g(w) + h(x.v),\\
&= g(x) + c(w.o) + h(n.v) - c(x.o),\\
&= f_L(x) - c(x.o) + c(w.o),\\
&=f(n)+c(w.o),\\
\end{align}
so in particular, $f_L(w) < f_L(y)$ because $f(n)+c(w.o) < f(n)+c(y.o)$.  Again suppose that
$w$ makes full progress towards the goal (the pessimism was entirely
unwarranted), so $h(w.v) = h(x.v)-c(w.o)$.  So any of its cheap
children, say $z$, satisfies:
\begin{align}
f_L(z) &= g(w) + c(z.o) + h(x.v) - c(w.o),\\
&= f_L(w) - c(w.o) + c(z.o),\\
&= f_L(x) - c(x.o) + c(w.o) - c(w.o) + c(z.o),\\
&= f_L(x) - c(x.o) + c(z.o),\\
&= f(n)+c(z.o).
\end{align}
Inductively, any low-cost-reachable descendant, say $x'$, that makes full
heuristic progress, has an $f_L$ value of the form $f(n) + c(x'.o)$, and in particular, $f_L(x') < f_L(y)$, that is, all such
descendants are expanded prior to $y$.  Generalizing, any low-cost-reachable and not
heuristically bad descendant of $x$ is expanded prior to $y$ (where
the bound on heuristic badness is $c(y.o)$ --- the amount by which $y$
is pessimistically considered heuristically bad).  Once
$y$ itself is finally expanded, then its descendants can compete with
the descendants of $x$ on even footing, so in particular some of the expensive
exits of the low-cost subspace underneath $y$ may very well be
explored prior to some of the expensive (heuristically or immediately)
exits of the low-cost subspace underneath $x$ --- in contrast with the
low-cost subspaces themselves, which were explored in depth-first
fashion, i.e., all of $x$'s subspace before all of $y$'s subspace.

Then LAMA exhibits a curious, temporary,
depth-first behavior initially, but in the large exhibits the normal
breadth-first bias of systematic search.  Depth-first behavior
certainly results in finding an increasingly good sequence of plans
to the same state:  At every point in the best plan to some state where a less-expensive
sibling leads to a slightly worse plan to the same state is a point at
which depth-first behavior finds worse plans first. The travel domain is very
strongly connected, so there are many such opportunities.

\noindent{\bf Overhead.} Consider two paths to the same
plane-shuffling state, the second one actually (but not heuristically)
better.  Then LAMA has already expanded the vast majority, if not the
entirety, of the associated passenger-shuffling subspace before
finding the second plan. \Ignore{(Well, as much as it can reach using
  just preferred operators, supposing that technique is enabled, but
  the point stands.)}  That entire set is then re-expanded.  The
size-based approach is not compelled to exhaust the
passenger-shuffling subspaces in the first place (indeed, it is
compelled to backtrack to other possibilities), and so in the same
situation ends up performing less re-expansion work within each
passenger-shuffling subspace.  Then even if the size-based approach is
overall making more mistakes in its use of planes (finding worse plans
first), which is to be expected, the price per such mistake is notably
smaller.

\Ignore{If such depth-first bias is still present at heuristic weight 1, then
clearly the difficulty will only be greater at heuristic weight 5 (or infinity).  So on
larger problems, it is easy to see that the cost-based search will be
stuck re-expanding all the same states many times over if the
heuristically preferred solution has already been reported (and so is
pruned upon re-discovering it, leading to a large amount of
backtracking and re-expansion in the neighborhood of that solution).
Indeed, turning preferred operators off can lead to better overall
performance (worse initial solutions found later in time, but, better
final solutions before running out of time/memory), a fact well
explained by the hypothesis that LAMA is stuck in the neighborhood of
a particular, pruned, solution.  Similarly, removing the additional
open list on landmarks worsens performance significantly (the
additional open list has the effect of increasing breadth of exploration).
As a final piece of evidence, LAMA-size runs out of memory sooner (on the large problems),
indicating that it is spending a greater fraction of its time
expanding new states.}

\begin{table}[ht]
\centerline{\begin{tabular}{|c||c|c|}
\hline
Domain & LAMA & LAMA-size\\
\hline
Rendezvous              & 70.8\% &83.0\%\\
Elevators               & 79.2\% &93.6\%\\
Woodworking             & 76.6\% &64.1\%\\
\hline
\end{tabular}}
\caption{IPC metric on LAMA variants.}
\end{table}

\noindent{\bf Results.}\footnote{New best plans
  for Elevators were found (largely by LAMA-size).  The baseline planner's score is 71.8\% against the better reference plans.}  With more than about 8 total passengers, LAMA
is unable to complete any search stage except the first (the greedy search). 
For the same problems, LAMA-size finds the same first plan (the
heuristic values differ, but not the structure), but is then subsequently able to
complete further stages of search.  In so doing it sees marked
improvement in cost; on the larger problems this is due only to
finding better variants on the greedy plan.  Other domains are
included for broader perspective, woodworking in particular was chosen
as a likely counter-example, as all the actions concern just one
type of physical object and the costs are not wildly different.  For
the same reasons we would expect LAMA to out-perform LAMA-size in some
cost-enhanced version of Blocksworld.  For a comprehensive empirical
analysis, see~\cite{lama-journal}.

\noindent{\bf Summary.} LAMA is out-performed by LAMA-size, due to the former
spending far too much time expanding and re-expanding states in the $\eps$-cost
plateaus.  It fails in ``depth-first'' mode: finding
not-cheapest almost-solutions, exhausting the associated cheap subspace,
backtracking, finding a better path to the same state,
re-exhausting that subspace, \dots, in particular exhausting memory
extremely slowly (it spends all of its time re-exhausting the \emph{same}
subspaces).

\subsection{Analyzing the Performance of SapaReplan}

The contrasting failure mode, ``breadth-first'', is characterized by
exhausting each such subspace as soon as it is encountered, thereby
rapidly exhausting memory, without ever finding solutions.  This is
largely the behavior of SapaReplan (which does eager evaluation), with cost-based methods running
out of memory (much sooner than the deadline, 30 minutes) and
size-based methods running out of time.  So for SapaReplan it is the
size-based methods that are performing many more re-expansions, as in a
much greater amount of time they are failing to run out of memory.  
From the results, these re-expansions must be in a useful area of the
search space.  

In particular it seems that the cost-based methods must indeed be
exhausting the passenger-shuffling spaces more or less as soon as they
are encountered --- as otherwise it would be impossible to both
consume all of memory yet fail to find better solutions. (Even with
fuel there are simply too few distinct states modulo
passenger-shuffling.)  However, they do find solutions before getting trapped, in
contradiction with theory.  

The explanation is just that the cost-based methods are run with large (5)
heuristic weight, thereby introducing significant depth-first bias
(but not nearly so significant as with pessimistic delayed evaluation), so
that it is possible for them to find a solution before attempting to
exhaust such subspaces.  It follows that they find solutions within
seconds, and then spend minutes exhausting memory (and indeed that is
what occurs).  
The size-based methods are run with small heuristic weight (2) as they tend to perform
better in the long run that way.  It would be more
natural to use the same heuristic weight for both types, but, the
cost-based approaches do conform to theory with small heuristic
weights --- producing no solutions, hardly an interesting comparison.

\subsection{Summary}

Both planners are capable of finding solutions to problems in the
travel domain despite the use of a cost-based evaluation function by
using various tricks to induce a limited amount of depth-first
behavior in an $A^*$-framework.  This has the potential effect of
delaying exploration of the $\eps$-cost plateaus slightly, past the
discovery of a solution, but still each planner is ultimately trapped
by such plateaus before being able to find really good solutions.
Then such tricks are mostly serving to mask the problems of cost-based
search (and $\eps$-cost), as they merely delay failure by just enough
that one can imagine that the planner is now effective (because it
returns a solution where before it returned none).  Using a size-based
evaluation function more directly addresses the existence of cost
plateaus, and not surprisingly leads to improvement over the
equivalent cost-based approach --- even with LAMA.

\end{document}